\documentclass{article}
\usepackage{amsmath,graphicx,mlspconf}

\usepackage{amsmath}
\usepackage{graphicx,psfrag,epsf}
\usepackage{enumerate}
\usepackage{url} 
\usepackage{hyperref}
\usepackage{url} 
\usepackage{amssymb,amsfonts,mathtools}
\usepackage{bm}
\usepackage{lipsum}
\usepackage{color, colortbl}
\usepackage{enumitem}
\usepackage{etoolbox}
\usepackage{cite}
\usepackage{caption}
\usepackage{subfigure}
\usepackage{algorithm}
\usepackage{algorithmic}
\usepackage{balance}

\newtheorem{definition}{Definition}
\newtheorem{remark}{Remark}
\newtheorem{theorem}{Theorem}

\newtheorem{lemma}{Lemma} 

\newtheorem{assumption}{Assumption}

\usepackage{xspace}
\usepackage{bbm}
\usepackage{mathrsfs}

\newcommand{\Rc}{\mathcal{R}}
\newcommand{\Sc}{\mathcal{S}}

\newcommand{\Av}{{\bf A}}

\newcommand{\Gv}{{\bf G}}

\newcommand{\Iv}{{\bf I}}

\newcommand{\Xv}{{\bf X}}

\newcommand{\Wv}{{\bf W}}

\newcommand{\bv}{{\bf b}}

\newcommand{\ev}{{\bf e}}
\newcommand{\fv}{{\bf f}}
\newcommand{\gv}{{\bf g}}

\newcommand{\Ov}{{\bf O}}

\newcommand{\xv}{{\bf x}}
\newcommand{\yv}{{\bf y}}
\newcommand{\wv}{{\bf w}}

\newcommand{\sv}{{\bf s}}

\newcommand{\thetav}{\boldsymbol \theta}

\newcommand{\betav}{\boldsymbol \beta}

\newcommand{\Sigmav}{\boldsymbol \Sigma}

\def\d{\delta}
\def\e{\epsilon}
\def\eps{\epsilon}

\newcommand{\Bern}{\mathrm{Bern}}

\newcommand{\Norm}{\mathcal{N}}

\def\textiid{i.i.d.\@\xspace}
\newcommand\iid{\ifmmode\text{ i.i.d. } \else \textiid \fi}

\newcommand{\Real}{\mathbb{R}}

\title{Differentially Private Variable Selection \\via the Knockoff Filter}
   \name{Mehrdad Pournaderi and Yu Xiang}
   \address{University of Utah\\
                     50 Central Campus Dr 2110, Salt Lake City, USA\\
                     Email: \{mehrdad.pournaderi, yu.xiang\}@utah.edu}

\begin{document}

\maketitle

\begin{abstract}
The knockoff filter, recently developed by Barber and Cand{\`e}s, is an effective procedure to perform variable selection with a controlled false discovery rate (FDR). We propose a private version of the knockoff filter by incorporating Gaussian and Laplace mechanisms, and show that variable selection with controlled FDR can be achieved. Simulations demonstrate that our setting has reasonable statistical power.
\end{abstract}
\section{Introduction} \label{sec:intro}

A common theme in statistical inference and machine learning is to select a few relevant features or variables (among many) that are associated with the responses of interest. Controlling the expected proportion of falsely selected features (or the FDR~\cite{benjamini1995}) for the variable selection problem has attracted much attention since the development of the \emph{knockoff filter} procedure due to its flexibility (in constructing test statistics) as well as good finite-sample performances~\cite{barber2015controlling,barber2019knockoff}. In this procedure, a test statistic is computed for every feature of the model and the variables are selected by (data-dependent) thresholding the statistics according to the target FDR.
The knockoff filter has been generalized to various settings such as the high-dimensional case ($p>n$)~\cite{barber2019knockoff}, beyond linear models (the model-X framework)~\cite{candes2018panning}, and deep learning-based procedure~\cite{romano2019deep
}, with applications in a variety of domains~\cite{lu2018deeppink,fan2019ipad
}.
 
In this work, we aim to develop $(\e,\d)$-differentially-private statistics that ensure (finite-sample) FDR control while being empirically powerful. Differential privacy is currently the standard framework concerning the privacy of individuals. The differential privacy of other variable selection methods like the Lasso \cite{tibshirani1996regression}  and BH procedure \cite{benjamini1995} are investigated in \cite{thakurta2013differentially} and \cite{dwork2018differentially}, respectively. However, the privacy of the knockoff procedure hasn't been discussed in the literature yet. Related works are \cite{dwork2014analyze,sheffet2017differentially} where they consider the differential privacy of the sample covariance and the least-squares estimator. Although our approach is similar, the work is different in the sense that we take care of the FDR control and the knockoff design which is a function of the data. We adopt the sensitivity analysis and output perturbation method~\cite{dwork2006calibrating} and randomize the statistics by adding noise according to Gaussian and Laplace mechanism \cite{dwork2014algorithmic}.


\section{Preliminaries}
\label{sec:prob}

\subsection{Fixed-X Knockoff Filter}
Let $\yv\in \Real^n$ denote a vector of responses, and $\Xv\in\Real^{n\times p}$ a data matrix consisting of $p$ (explanatory) variables and $n$ samples. A canonical linear regression model is given by 
\begin{align}
	\yv = \Xv \betav + \wv,\label{linear_model}
\end{align}
where 
$\betav = [\beta_1,...,\beta_p]^\top\in \Real^p$ 
is the unknown regression vector and the model assumptions are as follows. 
\begin{assumption}
$\wv$ is Gaussian with constant variance $\sigma^2$ i.e. $\wv\sim \Norm(0, \sigma^2 \Iv_n)$. 
\end{assumption}

\begin{assumption}
$\Sigmav=\Xv^\top\Xv$ is invertible.
\end{assumption}
We say the $i$-th variable is a null variable if $\beta_i = 0$. The goal is to select the non-null variables (i.e. $\beta_i \neq 0$) without selecting too many nulls. 
For the linear model~\eqref{linear_model}, let $\hat{S}\subseteq\{1,2,\hdots,p\}$ denote the selected variables by some procedure.  Then the FDR is defined as follows
\begin{equation}
     \text{FDR}=\mathbb{E}\Bigg[\frac{\#\{j:\beta_j=0\ \text{and}\ j\in\hat{S}\}}{\#\{j: j\in\hat{S}\}\vee 1}\Bigg],
\end{equation}
where the expectation is taken over the randomness of the error terms, $\wv$. A selection rule controls the FDR at level $q$ if its corresponding FDR is guaranteed to be at most $q$. In the knockoff procedure, the FDR control is achieved by computing a vector of statistics $\Wv=(W_1, W_2,\hdots,W_p)^\top$ with components corresponding to each variable and selecting $\{j:W_j\geq T(q,\Omega)\}$, where the threshold $T\geq 0$ depends on the target FDR $q$ and the set of computed statistics, i.e., $\Omega=\{W_1,\hdots,W_p\}$. The main idea in computing the statistics is to use a fake copy of the design that preserve the correlation structure of $\Xv$ but destroys the relationship between $\Xv$ and $\yv$. For example, if we shuffle (or permute) the samples in $\Xv$, we preserve the geometry of $\Xv$ but the relation with $\yv$ would be diminished. The knockoff filter is based on a more elegant construction of such a fake design, namely, the knockoff design. 

\subsubsection{Knockoff Design}
 Knockoff copies are constructed using only the original features and the subtle geometrical relation between the data matrix $\Xv$ and its knockoff $\mathbf{\Tilde{X}}\in\Real^{n\times p}$ makes them an appropriate tool for FDR control. Let $\boldsymbol{\Sigma}=\Xv^\top\Xv$ denote the Gram matrix of the original features. The relationship between $\Xv$ and $\mathbf{\Tilde{X}}$ can be summarized as below
\begin{align}
    &\Gv := [ \Xv\ \mathbf{\Tilde{X}}]^\top[ \Xv\ \mathbf{\Tilde{X}}]\nonumber\\
    &\:\quad=\begin{bmatrix} {\boldsymbol{\Sigma}\qquad\qquad\quad \boldsymbol{\Sigma}-\text{diag}\{\sv\}}\\{\boldsymbol{\Sigma}-\text{diag}\{\sv\}\qquad\qquad\quad \boldsymbol{\Sigma}}\end{bmatrix},\label{eq:data_gram}
\end{align}
where $\sv\in\Real^p_+$ is chosen in a way that $\bf G$ is positive semidefinite. If $n\geq 2p$, $\mathbf{\Tilde{X}}$ can be constructed as follows
\begin{equation}
    \mathbf{\Tilde{X}}=\Xv\,({\bf I}_p-\boldsymbol{\Sigma}^{-1} \text{diag}\{\sv\})+\mathbf{\tilde{U}}{\bf C},\label{KOconst}
\end{equation}
where $\bf C$ is obtained by Cholesky decomposition of the Schur complement of $\Gv$ and $\mathbf{\tilde{U}}^{n\times p}$ is an orthonormal matrix that satisfies $\mathbf{\tilde{U}}^\top\Xv=0$. The existence of $\mathbf{\tilde{U}}$ is guaranteed by $n\geq 2p$. By looking at \eqref{eq:data_gram}, one can observe that the knockoff features matrix  $\tilde{\Xv}$ preserves the correlation structure of $\Xv$ and the vector $\sv$ gives us a degree of freedom to make knockoff features different than the original ones, e.g., if $\sv=0$ then $\mathbf{\Tilde{X}}=\Xv$. 
For maximum detection power, the knockoff and original features should be as orthogonal to each other as possible. The restriction on this matter is imposed by $\Gv\succeq 0$ which is the necessary and sufficient condition for $\mathbf{\Tilde{X}}$ to exist. 
\begin{remark}
 A suggestion in \cite{barber2015controlling} regarding the choice of $\sv$ for a normalized design, is to set $\text{diag}\{\sv\} =\text{min}( 2\lambda_{\text{min}}(\Sigmav),1)\,\Iv_p$. If $2\lambda_{\text{min}}(\Sigmav) \leq 1$, this choice (as we shall see in Lemma \ref{ev}) forces the minimum eigenvalue of $\Gv$ to be zero. For the differential privacy purposes we recommend using $\text{diag}\{\sv\} = \lambda_{\text{min}}(\Sigmav)\,\Iv_p$ to make the regression operators stable.
\end{remark}
\subsubsection{Statistics}
Using the knockoff features, we are now able to construct a vector of statistics $\Wv$ allowing FDR control. For instance, the Lasso penalized least-squares estimate can be used to construct the statistics.
\begin{equation}
    \boldsymbol{\hat{\beta}}^*:=\underset{\bv\,\in\Real^{2p}}{\arg\min}\Big\{\,\Big\|\yv-[\Xv\ \mathbf{\Tilde{X}}]\bv\Big\|_2^2\,+\lambda \|\bv\|_1 \Big\},\label{eq:LS_model}
\end{equation}
where $\lambda \geq 0$. Using this estimate, one way to define the statistics $\Wv$ is as follows
\begin{equation}
     W_i^{\textsc{LCD}} = |\hat{\beta}^*_i| - |\hat{\beta}^*_{i+p}|\ ,\qquad 1\leq i\leq p\ .\label{stat:LCD}
\end{equation}
We can also define the statistics differently,
\begin{equation}
    W_i = \text{sgn}(|\hat{\beta}^*_i| - |\hat{\beta}^*_{i+p}|)\,\max(|\hat{\beta}^*_i|,|\hat{\beta}^*_{i+p}|)\ .\label{lcsm}
\end{equation}
The key feature of these statistics that make them suitable for the purpose of FDR control is the \textit{\iid sign property} of the nulls, i.e., the signs of null statistics (signs of $W=\{W_j: \beta_j=0\}$) are independent of their magnitudes and have \iid $\Bern(1/2)$ distribution. We will look at this property in more details in the subsequent subsection. Unlike the nulls, non-null statistics are expected to get large positive values. Thus, a selection rule of the form $\{j:W_j\geq T\}$ for some $T>0$ seems reasonable. However, $T$ should be characterized such that guarantees FDR control. 

\subsubsection{IID Sign Property of Nulls}
In general, satisfying the following two properties is sufficient to guarantee the \iid sign property of the null statistics. Let $\Xv^{(i)}$ denote the $i$-th column of $\Xv$.

\smallskip
\textit{\textbf{Property I} : Swapping $\Xv^{(i)}$ and $\mathbf{\Tilde{X}}^{(i)}$ for all $i\in F$ and any $F\subseteq \{1,2,\hdots,p\}$ has the effect of switching the signs of $\{W_i:i\in F\}$. This property is called \textbf{antisymmetry} in \cite{barber2015controlling}.} 
\smallskip

In the least-squares problem \eqref{eq:LS_model}, it can be observed that swapping $\Xv^{(i)}$ and $\mathbf{\Tilde{X}}^{(i)}$ for any $i\in \{1,2,\hdots,p\}$ will swap $\hat{\beta}^*_i$ and $\hat{\beta}^*_{i+p}$ and this will result in switching the sign of $W_i$ according to~\eqref{stat:LCD} and \eqref{lcsm}.
\smallskip

\textit{\textbf{Property II}: Let $\Wv_{\text{swap}(F)}$ denote the vector of statistics obtained when $\Xv^{(i)}$ and $\mathbf{\Tilde{X}}^{(i)}$ are swapped in the regression problem for all $i\in F$ where $F\subseteq \{j:\beta_j = 0\}$. Then $\Wv_{\text{swap}(F)}$ and $\Wv$ have the same distribution, i.e. $\Wv_{\text{swap}(F)} \overset{d}{=}\, \Wv$.} 
\smallskip

 In the least-squares problem \eqref{eq:LS_model}, it can be shown that the estimated coefficients for null variables and their corresponding knockoff variable are exchangeable which immediately implies \textbf{Property II}. 
 
Using the \iid sign property for the null statistics, it can be shown that the selection rule $\hat{\Sc}(T)=\{j:W_j\geq T\}$ controls the FDR at level $q\in[0,1]$, where
\begin{equation}
    T = \min\Big\{t \in \Psi:\frac{1+\#\{j: W_j \leq -t\}}{\#\{j: W_j\geq t\}\vee 1}\leq q\Big\},\label{threshold}
\end{equation}
if $\Psi =\{|W_i|:i=1,2,\hdots p\}\setminus\{0\}$ is not empty and $T=+\infty$ if $\Psi = \emptyset$. 

\subsection{Differential Privacy}
Differential privacy provides a rigorous formulation of individual privacy that allows for the development of (randomizing) mechanisms that release some information about the data without revealing too much about any individuals. In other words, the goal is to have a guarantee that the effect of each observation in the output is negligible.
\begin{definition}
Two databases (matrices) are called neighbors or adjacent if they differ only in a single observation (row). 
\end{definition}

\begin{definition}[Differential Privacy]
A randomized mechanism $\mathcal{M}$ with domain $\mathcal{D}$ and range $\mathcal{R}$ is $(\e,\d)$-differentially private if for all pairs of neighboring inputs $\Av,\Av'\in\mathcal{D}$ and all measurable subsets $\Sc\subset \Rc$, it holds that,
\begin{equation*}
    \mathbb{P}(\mathcal{M}(\Av)\in\Sc)\le e^{\e}\,\mathbb{P}(\mathcal{M}(\Av')\in\Sc)+\d\ .
\end{equation*}
\end{definition}

The algorithms we are going to propose are based on the \textit{output perturbation} mechanisms. 
In general, the class of additive noise mechanisms perturb the output by adding random noise proportional to the (differential) sensitivity of the output, that is, $\mathcal{M}(\Av) = f(\Av) + Z$, where $f$ is some (vector-valued) mapping and $Z$ denotes the noise. The classical Gaussian mechanism~\cite{dwork2014algorithmic} and Laplace mechanism~\cite{dwork2006calibrating} adopt $Z\sim \Norm(0,\kappa^2 I)$ and $Z_i\overset{\iid}{\sim}\text{Lap}(\eta)$, where $\kappa^2$ and $\eta$ are determined by the $\ell_2$ and $\ell_1$-sensitivity of $f$, respectively. 

\begin{definition}\label{l2def}
The $\ell_2$-sensitivity of a function $f:\mathcal{D}(f)\longrightarrow\Real^p$ is defined as $\Delta_2 f = \underset{\Av \& \Av' \texttt{adjacent}}{max} \|f(\Av) - f(\Av')\|_2$.
In case of a function with range in matrices we use the Frobenius norm $\|.\|_F$ in the definition.
\end{definition}

\begin{theorem}[\cite{dwork2014algorithmic}]
For any $\epsilon\in(0,1)$ and $\delta\in(0,1)$ the Gaussian mechanism with $\kappa^2>2\ln{(1.25/\delta)}(\frac{\Delta_2 f}{\epsilon})^2$ is $(\eps,\delta)$-differentially private.\label{thm:GM}
\end{theorem}

\begin{theorem}[\cite{dwork2006calibrating}]
Defining $\ell_1$-sensitivity similar to Definition \ref{l2def} $\Delta_1 f = \underset{\Av \& \Av' \texttt{adjacent}}{max} \|f(\Av) - f(\Av')\|_1$, adding \iid $\text{Lap}\big(\frac{\Delta_1 f}{\eps}\big)$ to the components of $f(\Av)$ is $(\eps,0)$-differentially private.\label{LapM}
\end{theorem}

\section{Main Results}
In this section we provide two methods for computing differential private statistic under the assumptions 1,2, and the following assumption.
\begin{assumption}
$B$ is an upper bound for the $\ell_2$-norm of the rows in $\Xv$ and a lower bound for the $\ell_2$-norm of the columns.
\end{assumption}
For the purpose of variable selection with knockoffs, it is natural to normalize $\Xv$ by columns. So in both methods, the normalization is considered as a part of the procedure. Also, we let $\text{diag}\{\sv\} = \lambda_{\text{min}}(\Sigmav')\,\Iv_p$ in creating knockoffs \eqref{KOconst}, where $\Sigmav' = \Xv'^\top\Xv'$ and $\Xv'$ denotes the design matrix $\Xv$ normalized by the $\ell_2$-norm of its columns.
\subsection{Method I}
In this method we consider statistics that depend on the data through the following form,
\begin{equation}
    \widehat{\Wv}_1 = \fv([\Xv'\ \widetilde{\Xv'}]^\top[\Xv'\ \widetilde{\Xv'}]+E,[\Xv'\ \widetilde{\Xv'}]^\top\yv+\ev)\ ,\label{mthd1}
\end{equation}
where $E$ and $\ev$ denote the additive noise terms ensuring the differential privacy. Define $C_{\text{min}}=\underset{1\leq i \leq p}{\text{min}}\big\|\Xv^{(i)}\big\|_2 $ and $\eta^2= \frac{B^2}{C_{\text{min}}^2-B^2}$, where $\Xv^{(i)}$ denotes the $i$-th column of $\Xv$. In the first argument of \eqref{mthd1}, $E\in\Real^{2p\times2p}$ is a random term of the following structure,
\begin{equation}
    E = \thetav_1 \begin{bmatrix}{\Ov_p\qquad\Iv_p}\\{\Iv_p\qquad \Ov_p}\end{bmatrix}+\begin{bmatrix}{1\quad 1}\\{1\quad 1}
\end{bmatrix}\otimes\thetav_2 \nonumber\ ,
\end{equation}
where $\thetav_1\sim\text{Lap}\big(\eta^2(1+ \lambda_{\text{min}}(\Sigmav'))/\eps_1\big)$ and  $\thetav_2\in\Real^{p\times p}$ is a symmetric matrix with zero diagonal entries such that the elements of the upper triangle are drawn i.i.d. from $\Norm(0,\kappa_1^2)$, where $\kappa_1^2 \geq \frac{2\ln{(1.25/\delta)}}{\eps_2^2}\eta^4(\sqrt{2}+\|\Sigmav'\|_F)^2$,
and each lower triangle entry is copied from its upper triangle
counterpart. 
Regarding the second argument of \eqref{mthd1}, $\ev$ is a vector with \iid $\Norm(0,\kappa_2^2)$ components with $\kappa_2^2>\frac{2\ln{(1.25/\delta_1)}}{\epsilon^2}\Delta_2\big([\Xv'\ \widetilde{\Xv'}]^\top\yv\big)^2$ computed according to the following $\ell_2$-sensitivity, 
\begin{align}
    \Delta_2\big([\Xv'\ \widetilde{\Xv'}]^\top&\yv\big) =\, \label{mthd1_Sen}  \zeta^{\frac{1}{2}}\Big(2\sqrt{\gamma}+\eta\,\big(3+2\lambda_{\text{max}}(\Sigmav')\\+\lambda_{\text{min}}(\Sigmav')\big)^{\frac{1}{2}}&\Big)+\big\|\betav\big\|\Bigg(\sqrt{2}\, \Big(\frac{\eta}{B}-\frac{1}{C_{\text{min}}}\Big)\big\|\Sigmav\big\|_F+2\,\eta B \nonumber\\
    +\big(C_{\text{min}}-\frac{B}{\eta}\big)&\lambda_{\text{min}}(\Sigmav') +\eta^2\big(\lambda_{\text{min}}(\Sigmav')+1\big)\sqrt{C_{\text{min}}^2+B^2}\Bigg)\nonumber\ , 
\end{align}
 where, $\zeta = \frac{2p\,\sigma^2}{1-\sqrt{\frac{2}{p}\ln(2/\delta_2)}}$, $\gamma=2\lambda_{\text{max}}(\Sigmav')-\lambda_{\text{min}}(\Sigmav')$, and $\delta_2>2e^{-p/2}$.
\begin{theorem}[Privacy]\label{M1prv}
Releasing $\widehat{\Wv}_1$ is $(\eps+\eps',\delta+\delta')$-differentially private for arbitrary $\fv$, where $\delta' = \delta_1+\delta_2$ and $\eps' = \eps_1+\eps_2$.
\end{theorem}

\begin{theorem}[FDR control]\label{M1fdr}
Let $F\subseteq\{1,...,p\}$ and $P_F$ denote a $2p\times 2p$ symmetric permutation matrix corresponding to swapping the $i$-th column (or row) with $(i+p)$-th for all $i\in F$. Suppose the operator $\fv(\cdot,\cdot):\mathcal{D}\longrightarrow\Real^p$ (with $\mathcal{D}\subseteq\Real^{2p\times2p}\times \Real^{2p}$) is antisymmetric in the sense that  for all $(\Av,\bv)\in\mathcal{D}$, computing $\fv(P_F^\top\Av P_F,P_F\bv)$ has the effect of switching the signs of the components of $\fv(\Av,\bv)$ corresponding to $F$. Then, applying the knockoff selection rule on $\widehat{\Wv}_1$ controls the FDR.
\end{theorem}
For instance, we observe that if $\fv(\Av,\bv)$ denotes the vector of statistics computed using the estimate $\Av^{-1}\bv$, then $\fv(\cdot,\cdot)$  satisfies the antisymmetry condition in Theorem \ref{M1fdr}, because we have $(P_F\Av P_F)^{-1}P_F\bv=P_F\Av^{-1}\bv$.
\subsection{Method II}
In this method, we add noise directly to the least-squares estimate $\boldsymbol{\hat{\beta}^{'}}$, i.e. \eqref{eq:LS_model} with normalized design and $\lambda = 0$.
 \begin{theorem}[Privacy]\label{M2prv}
 
Fix $\delta_2>2e^{-p/2}$. If $\lambda_{\text{min}}(\Sigmav')>\frac{\eta^2}{1-\eta^2}$, then by the Gaussian mechanism, releasing $\boldsymbol{\hat{\beta}^{'}}+\ev$ with noise variance $\kappa^2>\frac{2\ln{(1.25/\delta_1)}}{\epsilon^2} \Delta_2(\boldsymbol{\hat{\beta}^{'}})^2$ and the following $\ell_2$-sensitivity,

\begin{align}
    \Delta_2(\boldsymbol{\hat{\beta}^{'}}) = 
    \frac{2\,\zeta^\frac{1}{2}}{\sqrt{(1-\eta^2)\lambda_{\text{min}}(\boldsymbol{\Sigma^{'}})-\eta^2}} +\Big(C_{\text{min}}-\frac{B}{\eta}\Big) \big\|\betav\big\|\ ,\nonumber
\end{align}
is $(\eps,\delta_1+\delta_2)$-differentially private, where $\zeta = \frac{2p\,\sigma^2}{1-\sqrt{\frac{2}{p}\ln(2/\delta_2)}}$.\label{thm:olspert}
\end{theorem}

\begin{theorem}[FDR control]\label{M2fdr}
Let $F\subseteq\{1,...,p\}$ and define $P_F$ as in Theorem \ref{M1fdr}. Suppose the operator $\gv(\cdot):\mathcal{D}\longrightarrow\Real^p$ (with $\mathcal{D}\subseteq\Real^{2p}$) is antisymmetric in the sense that for all $\bv\in\mathcal{D}$, computing $\gv(P_F\bv)$ has the effect of switching the signs of the components of $\gv(\bv)$ corresponding to $F$. Then, for any vector $\ev$ with \iid elements, applying the knockoff selection rule on $\widehat{\Wv}_2=\gv(\boldsymbol{\hat{\beta}^{'}}+\ev)$ will control the FDR.
\end{theorem}

\begin{remark}
The sensitivities we compute depend on the data, so the additive noise is calibrated according to the observed data set. As a result, the privacy holds in a local sense and not globally. 
In Method II, the term $\lambda_{\text{min}}(\Sigmav')$ in the denumerator can make the $\ell_2$-sensitivity $\Delta_2(\boldsymbol{\hat{\beta}^{'}})$, sensitive to the data. This may be an issue in case the adversary has some side information about the data. However, we can use the Ridge regression to alleviate this issue and stabilize the regression operator, i.e., if we add $\omega^2\,\Iv_{2p}$ to $\Gv'$ we get $\lambda_{\text{min}}(\Gv')=\lambda_{\text{min}}(\Sigmav')+\omega^2$, which is lower bounded by $\omega^2$. 
Also, the sensitivities depend on the unknown parameters $\sigma^2$ and $\betav$ which should be estimated or bounded for the use in practice.
\end{remark}

\section{Simulation Results}
\label{sec:simulation}
In this section we present simulation results on synthetic data sets for both Method I and II. The sample size $n$ is swept between $10^3$ to $10^{6.6}$. We have $p = 50$ measurements per sample and the entries of the design matrix $\Xv$ are generated \iid according to $\Norm(0,\Iv_p)$. The responses $\yv$ are generated according to the linear model \eqref{linear_model} with noise variance $\sigma^2 = 1$ and $k=15$ (non-zero) underlying coefficients of the same amplitude $A=4.5$ and sign. We adopt the OLS estimator and the \textit{Coefficient Signed Max} (CSM) statistic \eqref{lcsm}. The target FDR is $q=0.2$ and the plots are based on averaging 250 trials. Each trial is $(0.2,\frac{2p}{n})$-differentially private, meaning we let the overall $\delta$ parameter drop as the sample size grows, so the privacy hold in a more strict sense for larger sample sizes\footnote{In practice it is common to take $\delta$ to scale as $\frac{1}{n}$.}. In Method I we set $\e = 0.1$, $\e_1=\e_2=0.05$, and $\d=\d_1=\d_2$. In Method II we have $\e = 0.2$ and $\d_1=\d_2$. The power is defined as follows,
\begin{equation}
     \text{Power}=\frac{1}{k}\,\mathbb{E}\big[\#\{j:\beta_j\neq 0\ \text{and}\ j\in\hat{S}\}\big]\ .
\end{equation}

\begin{figure}[h]
    \centering
    \includegraphics[width=0.9\linewidth]{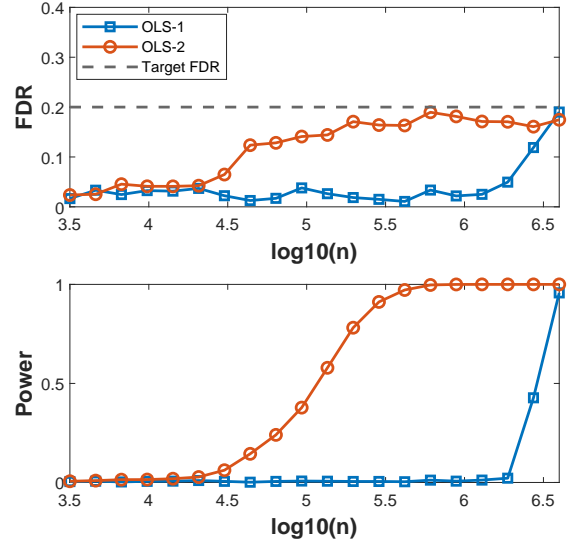}
    \caption{Power and FDR for the two differentially private knockoff procedures. The FDR is controlled at the desired level $q=0.2$ for both methods. However, Method II has the superior power in this experiment.}
    \label{fig:my_label}
\end{figure}



\appendix
\section{Appendix: Privacy Proofs}
In the differential privacy proofs we use the following lemma.
\begin{lemma}\label{ev}
If we set $\text{diag}\{\sv\} = \lambda_{\text{min}}(\Sigmav)\,\Iv_p$ in creating knockoff variables \eqref{KOconst}, then 
\begin{equation}
    \lambda_{\text{max}}(\Gv)=2\lambda_{\text{max}}(\Sigmav)-\lambda_{\text{min}}(\Sigmav)\ ,
\end{equation}
\begin{equation}
    \lambda_{\text{min}}(\Gv)=\lambda_{\text{min}}(\Sigmav)\ ,
\end{equation}
where $\Sigmav$ denotes the Gram matrix $\Xv^\top\Xv$.
\end{lemma}
\textbf{Proof:}
We note that $\Sigmav-\lambda\Iv_p$ and $\Sigmav-\lambda_{\text{min}}(\Sigmav)\,\Iv_p$ commute, therefore for the block matrix $\Gv-\lambda\Iv_{2p}$ we have \cite{silvester2000determinants},
\begin{align}
    \text{det}(\Gv-\lambda\Iv_{2p}) = \text{det}\big(2\Sigmav-&\lambda_{\text{min}}(\Sigmav)\,\Iv_p-\lambda\Iv_p\big)\nonumber\\
    &\text{det}\big(\lambda_{\text{min}}(\Sigmav)\,\Iv_p-\lambda\Iv_p\big)\ .
\end{align}
Hence, the set of the eigenvalues of $\Gv$ is the union of eigenvalues of $2\Sigmav-\lambda_{\text{min}}(\Sigmav)\,\Iv_p$ and $\lambda_{\text{min}}(\Sigmav)\,\Iv_p$. Therefore,
\begin{align}
     \lambda_{\text{max}}(\Gv)&= \text{max}\big(2\lambda_{\text{max}}(\Sigmav)-\lambda_{\text{min}}(\Sigmav),\lambda_{\text{min}}(\Sigmav)\big)\nonumber\\
     & = 2\lambda_{\text{max}}(\Sigmav)-\lambda_{\text{min}}(\Sigmav)\ ,\label{lammax}
\end{align}
\begin{align}
\lambda_{\text{min}}(\Gv)&= \text{min}\big(2\lambda_{\text{min}}(\Sigmav)-\lambda_{\text{min}}(\Sigmav),\lambda_{\text{min}}(\Sigmav)\big)\nonumber\\
     & = \lambda_{\text{min}}(\Sigmav)\ .  \label{lammin}     
\end{align}
\textit{ We observe that if we set $\text{diag}\{\sv\} = 2\lambda_{\text{min}}(\Sigmav)\,\Iv_p$, the same argument will result in $\lambda_{\text{min}}(\Gv)=0$.}
\subsection{Method I (Theorem \ref{M1prv})}
\textbf{Proof:}
 To begin with, we compute the sensitivities for $\Gv'=[\Xv'\ \widetilde{\Xv'}]^\top[\Xv'\ \widetilde{\Xv'}]$. If $\text{diag}\{\sv\} = \lambda_{\text{min}}(\Sigmav')\,\Iv_p$ then,
\begin{align}
    \Gv'
    =\begin{bmatrix} {\boldsymbol{\Sigma'}\quad \boldsymbol{\Sigma'}}\\{\boldsymbol{\Sigma'}\quad \boldsymbol{\Sigma'}}\end{bmatrix} - \begin{bmatrix}{\Ov_p\qquad\quad\lambda_{\text{min}}(\Sigmav')\,\Iv_p}\\{\lambda_{\text{min}}(\Sigmav')\,\Iv_p\qquad\quad \Ov_p}\end{bmatrix}\ .
\end{align}
Suppose $\Xv_1$ and $\Xv_2$ differ only in one observation. We denote this data point by $\xv_1^\top\in\Real^p$ in $\Xv_1$ which is replaced by $\xv_2^\top$ in $\Xv_2$. Let $\Xv_1'=\Xv_1 D_1$ and $\Xv_2'=\Xv_2 D_2$, where $D_1$ and $D_2$ are the (positive) diagonal matrices normalizing $\Xv_1$ and $\Xv_2$ by columns, respectively. In this case we have,
\begin{align}
\Gv_1'-\Gv_2' =   \begin{bmatrix} {1\quad 1}\\{1\quad 1}
\end{bmatrix}&\otimes(\Sigmav_1' -\Sigmav_2') \\&\!\!\!\!\!-\big(\lambda_{\text{min}}(\Sigmav_1')-\lambda_{\text{min}}(\Sigmav_2')\big) \begin{bmatrix}{\Ov_p\qquad\Iv_p}\\{\Iv_p\qquad \Ov_p}\end{bmatrix}\nonumber\ .  
\end{align}
By the triangle inequality we get,
\begin{align}
    \|\Sigmav_1' -\Sigmav_2'\|_F &\leq 
    \|D_2^\top(\Sigmav_1 -\Sigmav_2)D_2\|+\\
    &\qquad\qquad\|D^*(D_1^\top\Sigmav_1 D_1) D^* -D_1^\top \Sigmav_1 D_1\|\ ,\nonumber
\end{align}
where $D^* = D_2 D_1^{-1}$. Let $C_i = \big\|\Xv_1^{(i)}\big\|_2$. By definition, ${D_1}_{(ii)} = \frac{1}{C_i}$ for $1\leq i\leq p$.
Also, according to Assumption 3 we have $ \frac{1}{\sqrt{C_i^2+B^2}}\leq{D_2}_{(ii)}\leq \frac{1}{\sqrt{C_i^2-B^2}}$. Therefore, $\frac{C_i}{\sqrt{C_i^2+B^2}}\leq{D^*}_{(ii)}\leq \frac{C_i}{\sqrt{C_i^2-B^2}}$.
Let $C_{\text{min}}=\underset{1\leq i \leq p}{\text{min}}\big\|\Xv_1^{(i)}\big\|_2$, then we have
\begin{equation}
    \|\Sigmav_1' -\Sigmav_2'\|_F \leq \frac{\|\Sigmav_1 -\Sigmav_2\|+B^2\,\|\Sigmav_1'\|}{C_{\text{min}}^2-B^2}\ .
\end{equation}
We note, $\|\Sigmav_1 -\Sigmav_2\|_F = \|\xv_1^\top\xv_1 -\xv_2^\top\xv_2\|_F \leq \sqrt{2}B^2$.
Hence,
\begin{equation}
    \|\Sigmav_1' -\Sigmav_2'\|_F \leq \eta^2(\sqrt{2}+\|\Sigmav_1'\|),\label{frob}
\end{equation}
where $\eta^2 = \frac{B^2}{C_{\text{min}}^2-B^2}$. Regarding $\lambda_{\text{min}}(\Sigmav_1')-\lambda_{\text{min}}(\Sigmav_2')$, by the triangle inequality we have,
\begin{align}\label{lamsenbeg}
    \big|\lambda_{\text{min}}(\Sigmav_1')-\lambda_{\text{min}}(\Sigmav_2')&\big|\leq\\ \big|\lambda_{\text{min}}(D_2^\top\Sigmav_2 & D_2)-\lambda_{\text{min}}(D_2^\top\Sigmav_1 D_2)\big|+\nonumber\\  \big|\lambda_{\text{min}}(D_1^\top&\Sigmav_1 D_1)-\lambda_{\text{min}}(D^* D_1^\top\Sigmav_1 D_1 D^*)\big|\ .\nonumber
\end{align}
Weyl inequalities \cite{horn94} imply,
\begin{align}
    &|\lambda_{\text{min}}(D_2^\top\Sigmav_2 D_2)-\lambda_{\text{min}}(D_2^\top\Sigmav_1 D_2)|\leq \eta^2, \label{wineq1}\\
    &|\lambda_{\text{min}}(\Sigmav_1')-\lambda_{\text{min}}(D^*\Sigmav_1' D^*)|\leq \eta^2 \lambda_{\text{min}}(\Sigmav_1')\ .\label{wineq2}
\end{align}
Thus, by summing up \eqref{wineq1} and \eqref{wineq2} we get
\begin{equation}
    \big|\lambda_{\text{min}}(\Sigmav_1')-\lambda_{\text{min}}(\Sigmav_2')\big|\leq \eta^2\big(1+ \lambda_{\text{min}}(\Sigmav_1')\big)\ .\label{lambdasen}
\end{equation}
Therefore, $\Gv'$ can be $(\eps_1+\eps_2,\delta)$-differentially protected by adding a noise term of the following structure,
\begin{equation}
    E = \thetav_1 \begin{bmatrix}{\Ov_p\qquad\Iv_p}\\{\Iv_p\qquad \Ov_p}\end{bmatrix}+\begin{bmatrix}{1\quad 1}\\{1\quad 1}
\end{bmatrix}\otimes\thetav_2 \nonumber\ ,
\end{equation}
where $\thetav_1$ and $\thetav_2$ are determined by Laplace and Gaussian mechanisms (Theorem \ref{LapM} and \ref{thm:GM}), respectively. 

Regarding the feature-response products $[\Xv'\ \widetilde{\Xv'}]^\top\yv$, suppose $\yv_1$ is the vector of responses corresponding to $\Xv_1$ and $\yv_2$ is the vector of responses corresponding to $\Xv_2$, so they differ only in one element. Now we look at the $\ell_2$-sensitivity. By the triangle inequality we have,
\begin{align}
   \big \|[\Xv_1'\ \widetilde{\Xv_1'}]^\top\yv_1&-[\Xv_2'\ \widetilde{\Xv_2'}]^\top\yv_2\big\|_2 \leq \label{FRPsen}\\ &\big\|[\Xv_1'\ \widetilde{\Xv_1'}]^\top\yv_1-[\Xv_1'\ \widetilde{\Xv_1'}]^\top\Xv_1\betav\big\|\nonumber\\&+\big\|[\Xv_2'\ \widetilde{\Xv_2'}]^\top\yv_2-[\Xv_2'\ \widetilde{\Xv_2'}]^\top\Xv_2\betav\big\|\nonumber\\
    &+ \big\|\big([\Xv_1'\ \widetilde{\Xv_1'}]^\top\Xv_1-[\Xv_2'\ \widetilde{\Xv_2'}]^\top\Xv_2\big)\betav\big\|\nonumber\ .
\end{align}
We note $\|[\Xv_1'\ \widetilde{\Xv_1'}]^\top\yv_1-[\Xv_1'\ \widetilde{\Xv_1'}]^\top\Xv_1\betav\big\|\sim\Norm(0,\sigma^2\Gv_1')$. Therefore, from the concentration results for the norm of the Gaussian measure \cite{vershynin2018high,blum2016foundations} we obtain
\begin{align}\label{cineq1}
  \big\|[\Xv_1'\ \widetilde{\Xv_1'}]^\top\yv_1-[\Xv_1'\ \widetilde{\Xv_1'}]^\top\Xv_1\betav\big\|_2\leq\sqrt{\zeta\,\lambda_{\text{max}}(\Gv_1')}\ ,
\end{align}
with probability at least $1-\delta_2/2$, where $\zeta = \frac{2p\,\sigma^2}{1-\sqrt{\frac{2}{p}\ln(2/\delta_2)}}$. The same argument as \eqref{cineq1} holds for the second term of \eqref{FRPsen}. By adding up the first two terms of the bound in \eqref{FRPsen}, we get
\begin{align} \label{frpsen1}
    \zeta^{\frac{1}{2}}\Big(\sqrt{\lambda_{\text{max}}(\Gv_1')}&+\sqrt{\lambda_{\text{max}}(\Gv_2')}\Big) \leq \\ \zeta^{\frac{1}{2}}&\Big(2\sqrt{\gamma}+\eta\,\big(3+2\lambda_{\text{max}}(\Sigmav_1')+\lambda_{\text{min}}(\Sigmav_1')\big)^{\frac{1}{2}}\Big),\nonumber
\end{align}
where $\gamma=2\lambda_{\text{max}}(\Sigmav_1')-\lambda_{\text{min}}(\Sigmav_1')$ and the inequality follows from \eqref{lammax}, \eqref{lambdasen}, and the fact that the same argument as \eqref{lamsenbeg}-\eqref{lambdasen}, holds for $\lambda_{\text{max}}(\Sigmav')$ as well\footnote{We also use the inequality $(a+b)^\frac{1}{2}\leq a^\frac{1}{2}+b^\frac{1}{2}$ for $a,b\geq0$.}.
Regarding the third term of~\eqref{FRPsen} we have,
\begin{align}\label{frpsen2}
    \big\|\big([\Xv_1'\ \widetilde{\Xv_1'}]^\top\Xv_1-[\Xv_2'\ \widetilde{\Xv_2'}]^\top&\Xv_2\big)\betav\big\|_2\leq\\\Bigg\|\begin{bmatrix}D_1\Sigmav_1-D_2\Sigmav_2\\D_1\Sigmav_1-D_2\Sigmav_2\end{bmatrix}-\nonumber\\\big(\lambda_{\text{min}}(\Sigmav_1')D_1^{-1}-&\lambda_{\text{min}}(\Sigmav_2')D_2^{-1}\big)\begin{bmatrix}\Ov_p\\ \Iv_p\end{bmatrix}\Bigg\|\big\|\betav\big\|\nonumber
    \\\leq \Bigg[\sqrt{2}\, \Big(\frac{\eta}{B}-\frac{1}{C_{\text{min}}}\Big)\big\|\Sigmav_1\big\|+2&\,\eta B +
    \big(C_{\text{min}}-\frac{B}{\eta}\big)\lambda_{\text{min}}(\Sigmav_1')\nonumber\\ +\,\eta^2\big(\lambda_{\text{min}}(\Sigmav_1'&)+1\big)\sqrt{C_{\text{min}}^2+B^2}\Bigg]\big\|\betav\big\|\ ,\nonumber
\end{align}
where the last inequality is obtained by using the triangle inequality via $D_2\Sigmav_1$ and $\lambda_{\text{min}}(\Sigmav_1')D_2^{-1}$.
Thus, the overall $\ell_2$-sensitivity can be computed by adding up the bounds in \eqref{frpsen1} and \eqref{frpsen2}, which results in the bound \eqref{mthd1_Sen}. 
The total bound holds with probability at least $1-\delta_2$. Therefore, $[\Xv\ \Tilde{\Xv}]^\top\yv$ can be $(\eps,\delta_1+\delta_2)$-differentially protected by adding a noise vector $\ev\sim\Norm(0,\kappa_2^2)$, where $\kappa_2$ is determined by Gaussian mechanism (Theorem \ref{thm:GM}).

\subsection{Method II (Theorem \ref{M2prv})}
\textbf{Proof:}
Let $\overline{\betav}_1=\begin{pmatrix}D_1^{-1}\betav\\\Ov_{p\times1}\end{pmatrix}$ and $\boldsymbol{\hat\beta_1^{'}}$ denote the OLS estimate based on the data set $[\Xv_1', \yv_1]$. By the triangle inequality,
\begin{equation}
    \big\|\boldsymbol{\hat\beta_1^{'}}-\boldsymbol{\hat\beta_2^{'}}\big\|_2 \leq \big\|\boldsymbol{\hat\beta_1^{'}}-\mathbf{\overline{\betav}_1}\big\|+\big\|\boldsymbol{\hat\beta_2^{'}}-\mathbf{\overline{\betav}_2}\big\|+\big\|\mathbf{\overline{\betav}_1}-\mathbf{\overline{\betav}_2}\big\|\ .
\end{equation}
We note $\big(\boldsymbol{\hat\beta_1^{'}}-\mathbf{\overline{\betav}_1}\big)\sim\Norm(0,\sigma^2{\Gv_1'}^{-1})$. Therefore similar to \eqref{cineq1} we have $\big\|\boldsymbol{\hat\beta_1^{'}}-\mathbf{\overline{\betav}_1}\big\|\leq\sqrt{ \frac{\zeta}{\lambda_{\text{min}}(\Gv_1')}}$
with probability at least $1-\delta_2/2$. Also,
\begin{align}
    \big\|\mathbf{\overline{\betav}_1}-\mathbf{\overline{\betav}_2}\big\|\leq & \big\|D_2^{-1}-D_1^{-1}\big\|_2\big\|\betav\big\| \ ,\\
    \leq & \Big(C_{\text{min}}-\sqrt{C_{\text{min}}^2-B^2}\Big) \big\|\betav\big\|\ .\nonumber
\end{align}
Hence, if $\lambda_{\text{min}}(\Sigmav_1')>\frac{\eta^2}{1-\eta^2}$, then using \eqref{lammin} and \eqref{lambdasen} we obtain the $\ell_2$-sensitivity of the OLS estimate,
\begin{align}\label{olssen}
\big\|\boldsymbol{\hat\beta_1^{'}}-\boldsymbol{\hat\beta_2^{'}}\big\| \leq &
    \sqrt{\frac{\zeta}{\lambda_{\text{min}}(\Sigmav_1')}}+\sqrt{\frac{\zeta}{(1-\eta^2)\lambda_{\text{min}}(\Sigmav_1')-\eta^2}}\nonumber\\&\qquad+\Big(C_{\text{min}}-\sqrt{C_{\text{min}}^2-B^2}\Big) \big\|\betav\big\|\ .\nonumber
\end{align}


\section{Appendix: FDR Control Proofs}
\label{app:fdr}
\subsection{Method I (Theorem \ref{M1fdr})}
\textbf{Proof:}
According to the antisymmetry property of $\fv$,
\begin{align}
     \widehat{\Wv}_1^F = \fv\Big(P_F^\top\big(\Gv+E\big)P_F,P_F^\top\big([\Xv\ \Tilde{\Xv}]^\top\yv+\ev\big)\Big)
 \end{align}
switches the signs of the components of $\widehat{\Wv}_1$ corresponding to $i\in F$. From \cite{barber2015controlling} we know $P_F^\top\Gv P_F = \Gv$ and $P_F^\top[\Xv\ \Tilde{\Xv}]^\top\yv \overset{d}{=} [\Xv\ \Tilde{\Xv}]^\top\yv$ for all $F\subseteq\Sc_0$ where $\Sc_0$ denotes the set of null variables. To show the \iid sign property of the nulls, we note that $E$, $\ev$ and $\yv$ are mutually independent. We note that $P_F^\top E P_F \overset{d}{=} E$ and $P_F^\top\ev \overset{d}{=} \ev$ hold for all $F\subseteq\{1,\hdots,p\}$ since the entries that are swapped under $P_F$ are generated \iid. Therefore, $\widehat{\Wv}_1^F\overset{d}{=}\widehat{\Wv}_1$.  This concludes the \iid sign property and FDR control.

\subsection{Method II (Theorem \ref{M2fdr})}
\textbf{Proof:}
We note that from \cite{barber2015controlling} we have $P_F^\top\boldsymbol{\hat{\beta}^{'}} \overset{d}{=} \boldsymbol{\hat{\beta}^{'}}$ for all $F\subseteq\Sc_0$. We observe $P_F^\top\ev \overset{d}{=} \ev$ for all $F\subseteq\{1,\hdots,p\}$. Independence of $\boldsymbol{\hat{\beta}^{'}}$, and $\ev$ will immediately imply the \iid sign property of nulls.

\small
\balance
\bibliographystyle{IEEEbib}
\bibliography{refs}

\end{document}